\documentclass[lettersize,journal]{IEEEtran}
\usepackage{amsmath,amsfonts}
\usepackage{algorithmic}
\usepackage{algorithm}
\usepackage{array}
\usepackage[caption=false,font=normalsize,labelfont=sf,textfont=sf]{subfig}
\usepackage{textcomp}
\usepackage{stfloats}
\usepackage{url}
\usepackage{verbatim}
\usepackage{graphicx}
\usepackage{cite}
\usepackage{multirow}
\usepackage{color}
\def\rr#1{\textcolor{black}{#1}}
\def\rrr#1{\textcolor{black}{#1}}
\allowdisplaybreaks[4]

\begin{document}

\title{\rr{Integrating} Lattice-Free MMI into End-to-End \\ Speech Recognition}

\author{Jinchuan Tian, \IEEEmembership{Student Member, IEEE},
Jianwei Yu, Chao Weng, Yuexian Zou, \IEEEmembership{Senior Member, IEEE} \\ and Dong Yu, \IEEEmembership{Fellow, IEEE}
\thanks{Jinchuan Tian and Yuexian Zou are with the Advanced data and signal
processing laboratory, School of Electric and Computer Science, Peking
University, Shenzhen Graduate School, Shenzhen, China. This work was done when
Jinchuan Tian was an intern at Tencent AI Lab.}
\thanks{Jianwei Yu, Chao Weng and Dong Yu are with Tencent AI LAB. Jianwei Yu and Chao Weng are also with Tencent ASR Oteam.}
\thanks{Jianwei Yu and Yuexian Zou are the corresponding authors. (tomasyu@tencent.com; zouyx@pku.edu.cn)}
}

\markboth{Journal of \LaTeX\ Class Files,~Vol.~14, No.~8, August~2021}%
{Shell \MakeLowercase{\textit{et al.}}: A Sample Article Using IEEEtran.cls for IEEE Journals}


\maketitle

\begin{abstract}

In automatic speech recognition (ASR) research, discriminative criteria have achieved superior performance in DNN-HMM systems. 
Given this success, the adoption of discriminative criteria is promising to boost the performance of end-to-end (E2E) ASR systems.
With this motivation, previous works have introduced the minimum Bayesian risk (MBR, one of the discriminative criteria) into E2E ASR systems. 
However, the effectiveness and efficiency of the MBR-based methods are compromised:
the MBR criterion is only used in system training, which creates a mismatch between training and decoding;
the on-the-fly decoding process in MBR-based methods results in the need for pre-trained models and slow training speeds.
To this end, novel algorithms are proposed in this work to integrate another widely used discriminative criterion, lattice-free maximum mutual information (LF-MMI), into E2E ASR systems not only in the training stage but also in the decoding process.
The proposed LF-MMI training and decoding methods show their effectiveness on two widely used E2E frameworks: Attention-Based Encoder-Decoders (AEDs) and Neural Transducers (NTs).
Compared with MBR-based methods, the proposed LF-MMI method: maintains the consistency between training and decoding; eschews the on-the-fly decoding process; trains from randomly initialized models with superior training efficiency.
Experiments suggest that the LF-MMI method outperforms its MBR counterparts and consistently leads to statistically significant performance improvements on various frameworks and datasets from 30 hours to 14.3k hours. 
The proposed method achieves state-of-the-art (SOTA) results on Aishell-1 (CER 4.10\%) and Aishell-2 (CER 5.02\%) datasets. 
Code is released\footnote{https://github.com/jctian98/e2e\_lfmmi}.

\end{abstract}

\begin{IEEEkeywords}
Automatic speech recognition, discriminative training, sequential training, minimum Bayesian risk, maximum mutual information, end-to-end.
\end{IEEEkeywords}

\section{Introduction}
\IEEEPARstart{I}{n} recent research of automatic speech recognition (ASR), great progress has been made due to the advances in neural network architecture design\cite{noam,conformer, emformer, dfsmn, mocha} and end-to-end (E2E) frameworks \cite{las, lasctc, rnnt, rna, survey1, survey2, survey3}. 
\rr{Without the compulsory forced-alignment and external language model integration, the end-to-end systems are also becoming increasingly popular due to its compact working pipeline.}
Today, E2E ASR systems have achieved state-of-the-art results on a wide range of ASR tasks \cite{espnet_result, wenet}. 
Currently, attention-based encoder-decoders (AEDs) \cite{lasctc, las} and neural transducers (NTs) \cite{rnnt} are two of the most popular frameworks in E2E ASR. 
In general practice, training criteria like cross-entropy (CE), connectionist temporal classification (CTC) \cite{ctc} and transducer loss \cite{rnnt} are adopted in AED and NT systems. 
However, all of the three E2E criteria try to directly maximize the posterior of the transcription given acoustic features but never attempt to consider the competitive hypotheses and optimize the model discriminatively.

Given the success of discriminative training criteria (e.g., MPE \cite{hybrid_mpe,hybrid_mpe2, mpe3, mpe4}, sMBR \cite{hybrid_mbr,hybrid_mbr2} and MMI\cite{lfmmi_16,lfmmi_18, pychain, boost_mmi, mmi_hadian, mmi_semi, mmi_transfer}) in DNN-HMM systems, integrating these criteria into E2E ASR systems is promising to further advance the performance of E2E ASR systems.
Although several efforts \rr{\cite{rwth_fusion, mbr_las, mwer_las, mwer_rnnt, mbr_rnnt, ms_hat_mwer_lm, ms_rnnt_mwer_lm, gg_hat_mwer_rare, gg_hybrid_mwer_rare, gg_mwer_length}} have been spent on this field, most of these works \rr{\cite{mbr_las, mwer_las, mwer_rnnt, mbr_rnnt, ms_hat_mwer_lm, ms_rnnt_mwer_lm, gg_hat_mwer_rare, gg_hybrid_mwer_rare, gg_mwer_length}} focus on applying minimum Bayesian risk (MBR) discriminative training criterion to E2E ASR systems.
However, The effectiveness and efficiency of these MBR-based methods are compromised by several issues.
Firstly, all of these methods adopt the MBR criterion in system training but still use the maximize-a-posterior (MAP) paradigm during decoding. The mismatch between system training and decoding can lead to sub-optimal recognition performance. 
Secondly, the Bayesian risk objective is usually approximated based on the N-best hypothesis list and the corresponding posteriors, which is generated by the on-the-fly decoding process. This process, however, requires a pre-trained model for initialization and makes these methods in the two-stage style.
Thirdly, the on-the-fly decoding process is much more time-consuming, which results in the slow training speed of MBR-based methods\cite{rwth_fusion}. 

To this end, we propose to integrate lattice-free maximum mutual information \cite{lfmmi_16, lfmmi_18} (LF-MMI, another discriminative training criterion) into E2E ASR systems, specifically AEDs and NTs in this work. 
Unlike the methods aforementioned that only consider the training stage, the proposed method addresses the training and decoding stages consistently.
To be more detailed, the E2E ASR systems are optimized by both LF-MMI and other non-discriminative objective functions in training.
During decoding, evidence provided by LF-MMI is consistently used in either beam search or rescoring. 
In terms of beam search, MMI Prefix Score is proposed to evaluate partial hypotheses of AEDs while MMI Alignment Score is adopted to assess the hypotheses proposed by NTs.
In terms of rescoring, the N-best hypothesis list generated without LF-MMI is further rescored by the LF-MMI scores.
Compared with the MBR-based methods, our method 
(1) maintains consistent training and decoding under MAP paradigm; 
(2) eschews the sampling process, approximate the objective through efficient finite-state automaton algorithms and work from scratch; and
(3) maintains a much faster training speed than its MBR counterparts.

Experimental results suggest that adding LF-MMI as an additional criterion in training can improve the recognition performance. 
Moreover, integrating LF-MMI scores in the decoding stage can further improve the performance of both AED and NT systems. 
The best of our models achieves CER of 4.10\%  and 5.02\% on Aishell-1 test set and Aishell-2 test-ios set respectively, which, to the best of our knowledge, are the state-of-the-art results on these two datasets.
The proposed method is also applicable to tiny-scale and large-scale ASR tasks: up to 19.6\% and 9.9\% relative CER reductions are obtained on a 30 hours corpus and a 14.3k-hour corpus respectively. 

To conclude, we propose a novel approach to integrate the discriminative training criteria, LF-MMI, into E2E ASR systems. 
The main contributions of this work are summarized as follow:
\begin{itemize}
    \item This paper is among the first works that propose to adopt the LF-MMI criterion in both training and decoding stages of AED and NT frameworks while previous works\cite{mbr_las, mwer_las, mwer_rnnt, mbr_rnnt} only consider the training process and only address a single E2E ASR framework. 
    \item This paper proposes an efficient way to incorporate discriminative training criteria into E2E system training. Compared with previous work\cite{mbr_las, mwer_las, mwer_rnnt, mbr_rnnt}, the proposed approach is free from the pre-trained model and the on-the-fly decoding process.  
    \item Three novel decoding algorithms with LF-MMI criterion are presented, covering the first-pass decoding and the second-pass rescoring of AED and NT systems.
    \item A systemic analysis between the MBR-based methods and proposed LF-MMI method is delivered to provide more insight into discriminative training of E2E ASR.
    \item The proposed method achieves consistent error reduction on various data volume and achieve state-of-the-art (SOTA) results on two widely used Mandarin datasets (Aishell-1 and Aishell-2).
\end{itemize}


\rr{This journal paper is an extension of our conference paper\cite{tian2021consistent}. The extension of this paper mainly falls in three ways: (1) the decoding algorithms, including the efficient computation of MMI Prefix Score and the look-ahead mechanism in MMI Alignment Score, are updated from the original version to achieve better performance; (2) the MBR-based methods are carefully investigated and compared with the proposed method, which is also untouched in the conference paper; (3) more detailed experimental analysis on various hyper-parameters and corpus scale (range from 30 hours to 14.3k hours) is provided while only preliminary results are reported in the conference paper.}
The rest of this paper is organized as follows. 
The MBR-based training methods in E2E ASR are analyzed in section \ref{mbr_theory}. 
The proposed LF-MMI training and decoding methods are presented in section \ref{mmi_training_theory} and section \ref{mmi_decoding_theory} respectively. 
The experimental setup and the results are reported in section \ref{exp_setup} and \ref{exp_result}. 
We discuss and conclude in section \ref{conclusion}. 

\section{MBR-based methods in E2E ASR}
\label{mbr_theory}
In the rest of this paper, $\mathbf{O}=[o_1, ..., o_T]$ and $\mathbf{W}=[w_1, ..., w_U]$ represent the input feature sequence and the token sequence with length $T$ and $U$ respectively. 
Here $o_t \in \mathbb{R}^d$ is a $d$-dimensional speech feature vector while $w_u \in \mathcal{V}$ is a token in a known vocabulary $\mathcal{V}$. 
In addition, we note $\textbf{O}_1^t=[o_1, ..., o_t]$ and $\textbf{W}_1^u=[w_1, ..., w_u]$ as the first $t$ frames and the first $u$ tokens of $\textbf{O}$ and $\textbf{W}$ respectively. 
Subsequently, $H(\textbf{W}_1^u)$ is defined as the set of token sequences that start with $\textbf{W}_1^u$. 
Finally, we define \textit{\textless sos\textgreater}  and \textit{\textless eos\textgreater}  as the \textit{start-of-sentence} and \textit{end-of-sentence} respectively while \textit{\textless blk\textgreater}  stands for the \textit{blank} symbol used in NT systems. 

In this section, we briefly introduce the MBR-based methods and their deficiencies.

\subsection{MBR-based methods in E2E ASR}
\label{mbr_survey}
The motivation of MBR-based methods in E2E ASR is to solve the mismatch between the training objective and the evaluation metric. Conventional E2E ASR models are optimized to maximize the posterior of the transcription given the acoustic features:
\begin{equation}
\label{map_train}
    \mathbf{J}_{\tt{MAP}}(\mathbf{W}, \mathbf{O}) = P(\mathbf{W}|\mathbf{O})
\end{equation}
where $\mathbf{W}$ is the transcription text sequence. In decoding, the most probable hypothesis is considered as the recognition result.
\begin{equation}
\label{map_decode}
    \hat{\mathbf{W}} = \mathop{\arg\max}_{\mathbf{W}\in \mathbf{H}([<sos>])}P(\mathbf{W}|\mathbf{O})
\end{equation}
However, as the \textit{edit-distance} (a.k.a., Word Error Rate, WER, for languages like English; Character Error Rate, CER, for languages like Chinese) between the hypothesis $\hat{\mathbf{W}}$ and the transcription $\mathbf{W}$ is usually adopted as the evaluation metric of ASR task\cite{horifst}, there is no guarantee that the hypothesis with the highest posterior is exactly the hypothesis with smallest \textit{edit-distance}, which is a mismatch problem and may result in sub-optimal system performance.

To alleviate this mismatch, the MBR-based methods have been proposed to directly take the \textit{edit-distance} as the training objective. Formally, the Bayesian risk objective to minimize is formulated as:
\begin{equation}
\begin{aligned}
\mathbf{J}_{\tt{MBR}}(\mathbf{W}, \mathbf{O}) =& \mathbb{E}_{P(\bar{\mathbf{W}}|\mathbf{O})}[u(\bar{\mathbf{W}}, \mathbf{W})] \\
                      =&  \sum_{\bar{\mathbf{W}} \in H([<sos>])} P(\bar{\mathbf{W}}|\mathbf{O}) \cdot u(\bar{\mathbf{W}}, \mathbf{W})
\end{aligned}
\label{eq_mbr}
\end{equation}
where the $u(\cdot)$ is the risk function (also known as utility function) while $\bar{\mathbf{W}}$ is any possible hypothesis. 
In most of the MBR-based methods in E2E ASR, \textit{edit-distance(·)} is adopted as the risk function so the objective to minimize is exactly the expected \textit{edit-distance}.
If so, the MBR methods are also known as minimum word error rate (MWER) methods.

\begin{figure}[htpb]
    \centering
    \includegraphics[width=8.5cm]{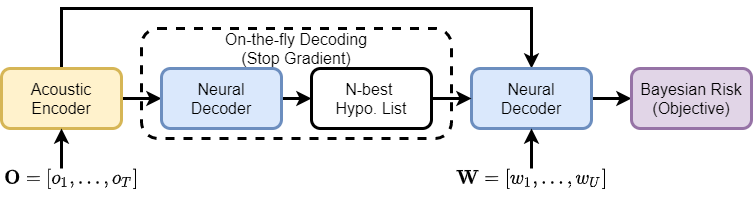}
    \caption{The workflow of MBR-based training methods. On-the-fly decoding is required to generate an N-best hypothesis list to approximate Bayesian risk.} 
    \label{mbr_fig}
\end{figure}

In real practice, however, the computation of the Bayesian risk is prohibitive, as the elements in the distribution $P(\bar{\mathbf{W}}|\mathbf{O})$ cannot be fully explored. 
Therefore, the objective of MBR-based methods is approximated by sampling over the set of all hypothesis:
\begin{equation}
    \mathbf{\hat{J}}_{\tt{MBR}}(\mathbf{W}, \mathbf{O}) = \frac{\sum_{\bar{\mathbf{W}}\in \mathcal{W}}P(\bar{\mathbf{W}}|\mathbf{O})\cdot u(\mathbf{\bar{W}}, \mathbf{W})} 
    {\sum_{\mathbf{W'}\in \mathcal{W}}P(\mathbf{W'}|\mathbf{O}) + \epsilon}
    \label{mbr_appro}
\end{equation}
where $\mathcal{W}$ is a subset of $H([\textless sos\textgreater])$ sampled from the posterior distribution $P(\bar{\mathbf{W}}|\mathbf{O})$ while the probability of each hypothesis in $\mathcal{W}$ is normalized by the summed probability over $\mathcal{W}$. $\epsilon$ is a smooth constant to avoid numerical errors. 
In existing MBR-based E2E methods, $\mathcal{W}$ is the N-best hypotheses list so the accumulated probability mass over $\mathcal{W}$ is maximized and a better approximation of $H([\textless sos\textgreater])$ is achieved. 
The N-best hypothesis list is generated by the decoding process, which is usually implemented on-the-fly during training (a.k.a, on-the-fly decoding). The workflow of MBR-based training methods is described in Fig.\ref{mbr_fig}.

Although all methods \cite{mbr_las, mwer_las, mwer_rnnt, mbr_rnnt, ms_hat_mwer_lm, ms_rnnt_mwer_lm, gg_hat_mwer_rare, gg_hybrid_mwer_rare, gg_mwer_length} roughly follow the paradigm above, there are still some differences among these methods. 
Firstly, \cite{mbr_las, mwer_las} work on AED systems while \cite{mwer_rnnt, mbr_rnnt, ms_hat_mwer_lm, ms_rnnt_mwer_lm, gg_hat_mwer_rare, gg_hybrid_mwer_rare, gg_mwer_length} serve NT systems and their derivatives\cite{ms_hat_mwer_lm,gg_hat_mwer_rare}. 
Secondly, \cite{mwer_las} also tries to generate $\mathcal{W}$ by random sampling rather than the N-best hypothesis list but performance degradation is observed. 
Thirdly, \cite{mwer_rnnt} proposes to replace the on-the-fly decoding with offline decoding and update the decoding results partially every after an epoch of data is exhausted for faster training speed. 
\rr{
Fourthly, \cite{mbr_rnnt, ms_hat_mwer_lm, ms_rnnt_mwer_lm, gg_hat_mwer_rare, gg_hybrid_mwer_rare} propose to integrate external language models during the MBR training to better leverage the linguistic information\cite{mbr_rnnt, ms_hat_mwer_lm, ms_rnnt_mwer_lm} or emphasize the rare words\cite{gg_hat_mwer_rare, gg_hybrid_mwer_rare}.
Fifthly, \cite{gg_mwer_length} discusses the impact of the input length in the MBR training.
}

Besides, there are also some attempts to introduce the MBR criterion during the decoding stage of DNN-HMM system \cite{mbr_decode, mbr_decode2} or in machine translation community\cite{coling_mbr}, which is out of the scope of this work. 

\subsection{Deficiencies of MBR-based methods}
\label{mbr_def}
Although MBR-based methods are found effective in improving recognition accuracy, we claim there are still several deficiencies in these methods.

\subsubsection{Mismatch between training and decoding}
As shown in Eq.\ref{map_train} and Eq.\ref{map_decode}, E2E ASR systems that adopt non-discriminative training criteria usually achieve the consistency between training and decoding: the posterior probability of the transcription is maximized during training while the goal of decoding is to find the hypothesis with maximum posterior probability. 
This consistency, however, fails in MBR-based methods, as the training objective is shifted to minimize the Bayesian risk (a.k.a, expected WER) but the searching target is still to find the most probable hypothesis, rather than the hypothesis with the smallest Bayesian risk or its approximation.



\subsubsection{The need for a pre-trained model for initialization}
As mentioned above, the approximation of the Bayesian risk objective in existing MBR-based methods requires the on-the-fly decoding process. In addition, the hypotheses generated on-the-fly should be roughly correct and occupy considerable probability mass to make the objective optimizable. For this reason, the MBR-based methods need a seed model pre-trained from non-discriminative training criteria for initialization to ensure the success of the on-the-fly decoding, which is in the two-stage style and leads to complex training workflow.

\subsubsection{Slow training speed}
Even though the on-the-fly decoding is successfully implemented, it would significantly slow down the training speed of the MBR-based methods multiple times. This is also reported in \cite{rwth_fusion}.

Besides the deficiencies aforementioned, we experimentally find the Bayesian risk objective is hard to be optimized and the performance ceiling of these methods is limited. As the training data is well-fitted after several epochs of training, it is much unlikely that an erroneous hypothesis with high probability mass would be generated during the on-the-fly decoding over the training data. Instead, we experimentally find that the transcription $\mathbf{W}$ with zero Bayesian risk usually occupies a dominant share of probability mass so the Bayesian risk is too small to optimize. We further discuss this observation in section \ref{exp_mbr2}.

\section{LF-MMI training for E2E ASR}
\label{mmi_training_theory}
This section briefly introduces the LF-MMI criterion in section \ref{intro_lfmmi}. Then the training methods with LF-MMI criterion are introduced in section \ref{train_aed} and section \ref{train_nt} for AED and NT systems respectively. 

\subsection{Lattice-Free Maximum Mutual Information criterion}
\label{intro_lfmmi}
As a discriminative training criterion, the Maximum Mutual Information (MMI) is used to discriminate the correct hypothesis from all hypotheses by maximizing the ratio as follows:
\begin{equation}
\begin{aligned}
    \mathbf{J}_{\tt{MMI}} &=\log P_{\tt{MMI}}(\mathbf{W}|\mathbf{O}) \\ &=\log\frac{P(\mathbf{O}|\mathbf{W})P(\mathbf{W})}{\sum_{\mathbf{\bar{W}}\in H([<sos>])}P(\mathbf{O}|\mathbf{\bar{W}})P(\mathbf{\bar{W}})}
\end{aligned}
\label{eq_mmi}
\end{equation}
where $\bar{\mathbf{W}}$ represents any possible hypothesis. 
Similar to MBR-based methods, directly enumerating $\bar{\textbf{W}}$ is almost impossible in practice. 
However, instead of using the N-best hypothesis list like MBR methods, Lattice-Free MMI\cite{lfmmi_16, lfmmi_18} is proposed to approximate the numerator and denominator in Eq.\ref{eq_mmi} by \textit{forward-backward} algorithm on two Finite-State Acceptors (FSAs). 
The log-posterior of $\mathbf{W}$ is then converted into the ratio of likelihood given the FSAs and acoustic features as follow\cite{pychain}:
\begin{equation}
\begin{aligned}
    \mathbf{J}_{\tt{LF-MMI}} = \log P_{\tt{LF-MMI}}(\mathbf{W}|\mathbf{O}) 
    \approx \log\frac{P(\mathbf{O}|\mathbb{G}_{num})}{P(\mathbf{O}|\mathbb{G}_{den})}
\end{aligned}
\label{eq_mmi_graph}
\end{equation}
where $\mathbb{G}_{num}$ and $\mathbb{G}_{den}$ denotes the FSA numerator graph and denominator graph respectively. 

Unlike the lattice-based MMI method\cite{boost_mmi}, the construction of the denominator graph in LF-MMI is identical to all utterances, which avoids the pre-decoding process before training and allows the system to be built from scratch. 
The mono-phone modeling units are adopted in the LF-MMI criterion, as a large number of modeling units (e.g. Chinese characters) makes the denominator graph computationally expensive and memory-consuming. 

\subsection{AED training with LF-MMI}
\label{train_aed}

\begin{figure}[htpb]
    \centering
    \includegraphics[width=8.5cm]{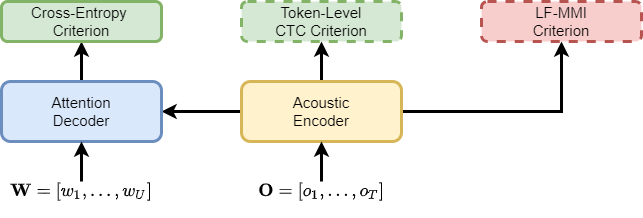}
    \caption{The diagram of the AED system architecture. Token-level CTC and LF-MMI criteria are adopted optionally} 
    \label{aed_fig}
\end{figure}

Attention-based Encoder-Decoders (AEDs) are a series of frameworks that adopt the encoder-decoder architecture to directly learn the mapping from the acoustic features to the transcriptions. 
In this work, we take the widely used hybrid CTC-attention framework\cite{lasctc} as the baseline model of AEDs. 
As shown in Fig.\ref{aed_fig}, the acoustic encoder is used to encode the acoustic features $\textbf{O}$.
Given the embeddings of the token sequence $\textbf{W}$ and the hidden output of the encoder, the attention decoder tries to predict the next tokens in auto-regressive style and is supervised by the cross-entropy criterion.
In \cite{lasctc}, the acoustic encoder is additionally supervised by the token-level CTC criterion\cite{ctc}.
Thus, the training objective of this system is the interpolated value between the cross-entropy (CE) criterion and token-level CTC criterion:
\begin{equation}
    \mathbf{J}_{\tt{AED}} = \alpha_{\tt{AED}} \cdot \mathbf{J}_{\tt{CE}} + (1-\alpha_{\tt{AED}}) \cdot \mathbf{J}_{\tt{CTC}}
\end{equation}
where $0<\alpha_{\tt{AED}}<1$ is a adjustable hyper-parameter. We refer \cite{lasctc} for more details.

Similary to \cite{lasctc}, the LF-MMI criterion is added as an auxiliary criterion to optimize the acoustic encoder in AED systems in this work. As shown in Eq.\ref{obj_aed}, the LF-MMI criterion is used to replace or cooperate with the CTC criterion.
\begin{equation}
    \label{obj_aed}
    \mathbf{J}_{\tt{AED}} = 
                 \left\{
                 \begin{aligned}
                 &\alpha_{\tt{AED}} \cdot \mathbf{J}_{\tt{CE}} + (1-\alpha_{\tt{AED}}) \cdot \mathbf{J}_{\tt{CTC}} + (1-\alpha_{\tt{AED}}) \cdot \mathbf{J}_{\tt{LF-MMI}}                                 \\
                 &\alpha_{\tt{AED}} \cdot \mathbf{J}_{\tt{CE}} + 
                 (1-\alpha_{\tt{AED}}) \cdot \mathbf{J}_{\tt{LF-MMI}}                                       \\
                 \end{aligned}
                 \right\} 
\end{equation}

\subsection{NT training with LF-MMI}
\label{train_nt}

\begin{figure}[htpb]
    \centering
    \includegraphics[width=8.5cm]{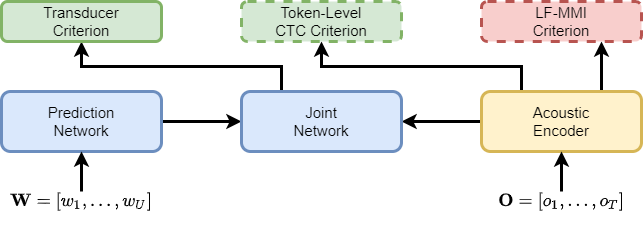}
    \caption{The diagram of the NT system architecture. Token-level CTC and LF-MMI criteria are adopted optionally} 
    \label{nt_fig}
\end{figure}
NT is another framework that is widely used in E2E ASR. As shown in Fig.\ref{nt_fig}, a typical NT system consists of an acoustic encoder, a prediction network and a joint network. Unlike the AED system that directly maximizes the posterior of $\mathbf{W}$ (a.k.a., $P(\mathbf{W}|\mathbf{O})$), the NT system tries to maximize the summed probability of a set of expanded hypotheses:
\begin{equation}
    \mathbf{J}_{\tt{NT}} = \log P(\mathbf{W}|\mathbf{O})
    = \log \sum_{\hat{\mathbf{W}} \in \mathcal{B}^{-1}(\mathbf{W})} P(\hat{\mathbf{W}}|\mathbf{O})
\end{equation}
where each element in $\mathcal{B}^{-1}(\mathbf{W})$, a.k.a, $\hat{\mathbf{W}}=[\hat{w}_1, ..., \hat{w}_{T+U}]$, is an expanded hypothesis generated from the transcription $\mathbf{W}$. The length of $\hat{\mathbf{W}}$ is $T+U$ and each element in $\hat{\mathbf{W}}$ is either a token in vocabulary $\mathcal{V}$ or $\textless blk\textgreater$. We refer \cite{rnnt} for more details.

Similar to the AED system, the LF-MMI criterion is added on the acoustic encoder and the token-level CTC criterion is optionally used. The training objective is then revised as:

\begin{equation}
    \label{obj_nt}
    \mathbf{J}_{\tt{NT}} = 
                 \left\{
                 \begin{aligned}
                 & \mathbf{J}_{\tt{NT}} + \alpha_{\tt{NT}} \cdot \mathbf{J}_{\tt{CTC}} + \alpha_{\tt{NT}} \cdot \mathbf{J}_{\tt{LF-MMI}}                                 \\
                 & \mathbf{J}_{\tt{NT}} + 
                 \alpha_{\tt{NT}} \cdot \mathbf{J}_{\tt{LF-MMI}}                                       \\
                 \end{aligned}
                 \right\} 
\end{equation}

\section{LF-MMI decoding for E2E ASR}
\label{mmi_decoding_theory}
To consistently use the LF-MMI criterion in both system training and decoding stages, three different decoding algorithms are proposed in this section. 
Specifically, MMI Prefix Score ($S^{\tt{pref}}_{\tt{MMI}}$) and MMI Alignment Score ($S^{\tt{ali}}_{\tt{MMI}}$) are proposed in section \ref{mps} and section \ref{mas} respectively to integrate LF-MMI scores into beam search of AEDs and NTs consistently. 
In addition, we also propose a rescoring method using LF-MMI in section \ref{rescore}. 
We finally compare our LF-MMI training and decoding method with MBR-based methods in section \ref{compare_mbr}.

\subsection{AED decoding with LF-MMI}
\label{mps}

Following Eq.\ref{map_decode}, given the acoustic feature sequence $\mathbf{O}$ and $\mathbf{W}_{1}^u=[\textless sos\textgreater]$, the goal of MAP decoding for AED systems is to search the most probable hypothesis $\hat{\mathbf{W}} \in \mathbf{H}([\textless sos\textgreater])$.
Normally, this searching process is approximated by the beam search algorithm. 
Assume $\mathbf{\Omega}_u$ is the set of active partial hypotheses with length $u$ during decoding. 
The set $\mathbf{\Omega}_{u}$ is recursively generated by expanding each partial hypothesis in $\mathbf{\Omega}_{u-1}$ and pruning those expanded partial hypotheses with lower scores. 
This iterative process would continue until a certain stopping condition is met. 
Typically, we set $\mathbf{\Omega}_0=\{[\textless sos\textgreater]\}$ while all hypotheses in any $\mathbf{\Omega}_u$ that end with $<eos>$ would be moved to a finished hypothesis set $\mathbf{\Omega}_F$ for final decision. 

The computation of partial scores is the core of beam search. Partial score $\alpha(\mathbf{W}_{1}^u, \mathbf{O})$ of a partial hypothesis $\mathbf{W}_{1}^u$ is recursively computed as:
\begin{equation}
    \alpha(\mathbf{W}_{1}^u, \mathbf{O}) = \alpha(\mathbf{W}_{1}^{u-1}, \mathbf{O}) + \log p(w_u|\mathbf{W}_{1}^{u-1}, \mathbf{O})
\end{equation}
where $\log p(w_u|\mathbf{W}_{1}^{u-1}, \mathbf{O})$ is the weighted sum of different log-probabilities possibly delivered by the attention decoder, the acoustic encoder and the language models. 
In this work, log-probability distribution provided by LF-MMI, namely $\log p_{\tt{LF-MMI}}(w_u|\mathbf{W}_{1}^{u-1}, \mathbf{O})$, is additionally considered as a component of $\log p(w_u|\mathbf{W}_{1}^{u-1}, \mathbf{O})$ with a pre-defined weight $\beta_{\tt{MMI}}$. 
The LF-MMI posterior $\log p_{\tt{LF-MMI}}(w_u|\mathbf{W}_{1}^{u-1}, \mathbf{O})$ can be derived from the first-order difference of $S^{\tt{pref}}_{\tt{MMI}}$: 
\begin{equation}
\label{eq_mmi_prefix_diff}
\log p_{\tt{LF-MMI}}(w_u|\mathbf{W}_{1}^{u-1}, \mathbf{O}) = S^{\tt{pref}}_{\tt{MMI}}(\mathbf{W}_{1}^{u},\mathbf{O}) - S^{\tt{pref}}_{\tt{MMI}}(\mathbf{W}_{1}^{u-1},\mathbf{O})
\end{equation}
where MMI Prefix Score $S^{\tt{pref}}_{\tt{MMI}}$ is defined as the summed probability of all hypotheses that start with the known prefix $\mathbf{W}_{1}^{u}$. 
\rr{
Given the fully known $\mathbf{W}$ and $\mathbf{O}$ (typically non-streaming scenarios), the MMI Prefix Score in Eq.\ref{eq_mmi_prefix} is firstly formulated. Secondly, $\mathbf{W}$ is split into two part with the known $u$. Next, the $\mathbf{O}$ is further split into two part. Since any $t$ is valid, the enumeration along t-axis is needed. In the approximation, we consider the conditional independence: for any known $t$, $\mathbf{O}_{1}^{t}$ is independent to $\mathbf{O}_{t+1}^{T}$ and $\mathbf{W}_{u+1}^{U}$ while $\mathbf{W}_{1}^{u}$ is independent to $\mathbf{O}_{t+1}^{T}$ and $\mathbf{W}_{u+1}^{U}$ (also see Eq.2.17 and Eq.2.18 in \cite{wang2015model}). In the fifth line, the probability sum over the set $\mathbf{H}(\mathbf{W}_{1}^{u})$ is equal to 1 and then discarded. In the final equation, each element $P_{\tt{MMI}}(\mathbf{W}_{1}^{u}|\mathbf{O}_{1}^{t})$ is approximated by Eq.\ref{eq_mmi_graph}, where $\mathbb{G}_{num}(\mathbf{W}_{1}^{u})$ is the numerator graph built from $\mathbf{W}_{1}^{u}$.
}


\rr{
\begin{align}
    \tiny
    \label{eq_mmi_prefix}
     & S^{\tt{pref}}_{\tt{MMI}}(\mathbf{W}_{1}^{u}, \mathbf{O}) = \log\sum_{\mathbf{W}\in \mathbf{H}(\mathbf{W}_{1}^{u})}P_{\tt{MMI}}(\mathbf{W}|\mathbf{O}) \notag \\
     =& \log \sum_{\mathbf{W}\in \mathbf{H}(\mathbf{W}_{1}^{u})} P_{\tt{MMI}}(\mathbf{W}_{u+1}^{U}|\mathbf{O}_{1}^{T})P_{\tt{MMI}}(\mathbf{W}_{1}^{u}|\mathbf{W}_{u+1}^{U}, \mathbf{O}_{1}^{T}) \notag \\
     =& \log\sum_{t=1}^{T}\sum_{\mathbf{W}\in \mathbf{H}(\mathbf{W}_{1}^{u})}  [P_{\tt{MMI}}(\mathbf{W}_{u+1}^{U}|\mathbf{O}_{1}^{t},\mathbf{O}_{t+1}^{T})\ \ \cdot \notag \\
     & \qquad \qquad \qquad \quad \ \  P_{\tt{MMI}}(\mathbf{W}_{1}^{u}|\mathbf{W}_{u+1}^{U}, \mathbf{O}_{1}^{t},\mathbf{O}_{t+1}^{T})] \notag \\
     \approx& \log\sum_{t=1}^{T}\sum_{\mathbf{W}\in \mathbf{H}(\mathbf{W}_{1}^{u})} P_{\tt{MMI}}(\mathbf{W}_{1}^{u}|\mathbf{O}_{1}^{t})P_{\tt{MMI}}(\mathbf{W}_{u+1}^{U}|\mathbf{O}_{t+1}^{T}) \notag \\
     =& \log\sum_{t=1}^{T}P_{\tt{MMI}}(\mathbf{W}_{1}^{u}|\mathbf{O}_{1}^{t}) \sum_{\mathbf{W}\in \mathbf{H}(\mathbf{W}_{1}^{u})}P_{\tt{MMI}}(\mathbf{W}_{u+1}^{U}|\mathbf{O}_{t+1}^{T}) \notag \\
     =& \log\sum_{t=1}^{T}P_{\tt{MMI}}(\mathbf{W}_{1}^{u}|\mathbf{O}_{1}^{t}) \approx \log\sum_{t=1}^{T}\frac{P(\mathbf{O}_{1}^{t}|\mathbb{G}_{num}(\mathbf{W}_{1}^{u}))}{P(\mathbf{O}_{1}^{t}|\mathbb{G}_{den})}
\end{align}}

In Eq.\ref{eq_mmi_prefix}, the accumulation of probability along the t-axis seems computationally expensive. 
However, several properties of it can be considered to greatly alleviate this problem. 
Firstly, unlike in the training stage, only the forward part of the \textit{forward-backward} algorithm is needed to calculate all terms in Eq.\ref{eq_mmi_prefix}. 
Secondly, the computation on the denominator graph is independent of the partial hypothesis $\mathbf{W}_{1}^{u}$, which can be done before the searching process and reused for any partial hypothesis proposed during beam search.
\rr{Thirdly, all items in series $P(O_1^t|G*)$ can be obtained through the computation of the last term $P(O_1^T|G*)$ by recording the intermediate variables of the forward process. As a result, the forward computation is conducted only once for each partial hypothesis. A more detailed description of this process is provided in Appendix A.}



\rr{Details of the AED beam search process with MMI Prefix Score integrated are shown in Algo. 1. In each decoding step, the log-posteriors from attention decoder, CTC and LF-MMI are updated into the score of each hypothesis $s(\mathbf{W_1^u})$ and the ended hypotheses will be moved into the finished hypothesis set. After the scores of all newly proposed hypotheses are updated, only $b$ hypotheses with highest scores can be kept in \textit{BeamPrune($\cdot$)} process before entering the next decoding step. The iteration will continue until the \textit{StopCondition}($\cdot$) is satisfied (see Eq. 50 in \cite{lasctc}). After the loop is ended, the hypotheses in the finished hypothesis set are sorted by their scores and returned.}

\begin{algorithm}[htpb]
\caption{AED beam search with MMI Prefix Score}
\begin{algorithmic}[1]
\label{algo_aed}
\REQUIRE acoustic feature sequence $\mathbf{O}$, vocabulary $\mathbf{V}$
\REQUIRE beam size $b$
\REQUIRE decoding weight $\beta_{\tt{ATT}}$, $\beta_{\tt{CTC}}$, $\beta_{\tt{MMI}}$
\STATE Initial Hypothesis Set $\Omega_0 \gets \{(\textless sos\textgreater, 0)\}$
\STATE Finished  Hypothesis Set $\Omega_F \gets \emptyset$ 
\STATE Decoding Step $u \gets 1$
\WHILE {not StopCondition($\Omega_F$, u)}

\STATE $\Omega_{u} \gets \emptyset$
\FOR {$(\mathbf{W}_1^{u-1}$, $s(\mathbf{W}_1^{u-1}))$ $\in \Omega_{u-1}$}
\FOR {$w_u\in\mathbf{V}$}
\STATE $\mathbf{W}_1^u \gets \mathbf{W}_1^{u-1} + w_u$
\STATE $s(\mathbf{W}_1^{u}) \gets s(\mathbf{W}_1^{u-1})$
\STATE $s(\mathbf{W}_1^{u}) \gets s(\mathbf{W}_1^{u}) + \beta_{\tt{ATT}} \cdot \log p_{\tt{ATT}}(w_u|\mathbf{W_1^{u-1}}, \mathbf{O})$ 
\STATE $s(\mathbf{W}_1^{u}) \gets s(\mathbf{W}_1^{u}) + \beta_{\tt{CTC}} \cdot \log p_{\tt{CTC}}(w_u|\mathbf{W_1^{u-1}}, \mathbf{O})$
\STATE $s(\mathbf{W}_1^{u}) \gets s(\mathbf{W}_1^{u}) + \beta_{\tt{MMI}} \cdot \log p_{\tt{MMI}}(w_u|\mathbf{W_1^{u-1}}, \mathbf{O})$
\ENDFOR

\IF {$w_u == \textless eos\textgreater$}
\STATE ${\Omega}_F\gets {\Omega}_F$ $\cup$ $(\mathbf{W}_1^{u}, s(\mathbf{W}_1^{u}))$ 
\ELSE
\STATE ${\Omega}_u\gets {\Omega}_u$ $\cup$ $(\mathbf{W}_1^{u}, s(\mathbf{W}_1^{u}))$ 
\ENDIF
\ENDFOR
\STATE $\Omega_u \gets$ BeamPrune($\Omega_u, b$) 
\STATE $u\gets u+1$
\ENDWHILE
\RETURN Sorted$(\Omega_F)$
\end{algorithmic}
\end{algorithm}

\subsection{NT decoding with LF-MMI}
\label{mas}
For NTs, MMI Alignment Score $S^{\tt{ali}}_{\tt{MMI}}$ is proposed to cooperate with the decoding algorithm ALSD\cite{alsd}. 
Note tuple ${(\textbf{W}_{1}^{u}, s_{t}(\textbf{W}_{1}^{u}))}$ as a hypothesis where $\textbf{W}_{1}^{u}$ is the output sequence (including no \textit{\textless blk\textgreater}) with length $u$, $s_{t}(\textbf{W}_{1}^{u})$ is the hypothesis score.
The subscript $t$ in $s_{t}(\textbf{W}_{1}^{u})$ means the hypothesis is aligned to first $t$ frames $\textbf{O}_{1}^{t}$. 

As hypotheses in NT decoding obtain explicit alignments $(t, u)$, the proposed $S^{\tt{ali}}_{\tt{MMI}}$ should also depend on this t-u relationship. Unlike the MMI Prefix Score that only considers the $u$-axis, we define the MMI Alignment Score as: 
\begin{equation}
    S^{\tt{ali}}_{\tt{MMI}}(\textbf{W}_{1}^{u}, \textbf{O}_{1}^{t}) = \mathop{\max}_{0\leq i \leq \tau} \log P_{\tt{MMI}}(\textbf{W}_{1}^{u}|\textbf{O}_{1}^{t+i})
\end{equation}
where $\tau$ is a hyper-parameter named \textit{look-ahead steps}. Thus, it is natural to conduct the beam search process of ALSD based on the interpolated value of the NT log-posterior and the MMI Alignment Score (also a log-posterior). In our implementation, once a new hypothesis is proposed (a $\textless blk\textgreater$ or a \rr{new} non-blank token is added), its score $s_{t+1}(\textbf{W}_{1}^{u})$ or $s_{t}(\textbf{W}_{1}^{u+1})$ is computed recursively using Eq.\ref{mas_blank} and Eq.\ref{mas_token}:

\begin{equation}
\label{mas_blank}
\begin{aligned}
     s_{t+1}(\textbf{W}_{1}^{u}) =& s_{t}(\textbf{W}_{1}^{u}) + \log p^{\tt{NT}}_{t+1}(\textless blk\textgreater|\mathbf{W_1^{u}}, \mathbf{O}) \\ 
     & +  \beta_{\tt{MMI}} * (S^{\tt{ali}}_{\tt{MMI}}(\textbf{W}_{1}^{u},\textbf{O}_{1}^{t+1}) - S^{\tt{ali}}_{\tt{MMI}}(\textbf{W}_{1}^{u},\textbf{O}_{1}^{t}))
\end{aligned}
\end{equation}

\begin{equation}
\setlength\abovedisplayskip{0cm}
\setlength\belowdisplayskip{0cm}
\label{mas_token}
\begin{aligned}
     s_{t}(\textbf{W}_{1}^{u+1}) =& s_{t}(\textbf{W}_{1}^{u}) + \log p^{\tt{NT}}_{t}(w_{u+1}|\mathbf{W_1^{u}}, \mathbf{O}) \\ 
     & +  \beta_{\tt{MMI}} * (S^{\tt{ali}}_{\tt{MMI}}(\textbf{W}_{1}^{u+1},\textbf{O}_{1}^{t}) - S^{\tt{ali}}_{\tt{MMI}}(\textbf{W}_{1}^{u},\textbf{O}_{1}^{t}))
\end{aligned}
\end{equation}
 where the score provided by MMI criterion is obtained from the first-order difference over $S^{\tt{ali}}_{\tt{MMI}}(\textbf{W}_{1}^{u},\textbf{O}_{1}^{t})$ along t-axis or u-axis respectively. $\beta_{\tt{MMI}}$ is an adjustable hyper-parameter in decoding stage. Similar to $S^{\tt{pref}}_{\tt{MMI}}$, we approximate $S^{\tt{ali}}_{\tt{MMI}}$ by Eq.\ref{eq_mmi_graph}. The denominator scores can also be reused in all hypotheses and the series $P_{\tt{MMI}}(\textbf{W}_{1}^{u}|\textbf{O}_{1}^{t+i}), i=0,...,\tau$ can also be computed in one go.

\begin{figure}
    \centering
    \includegraphics[width=8.5cm]{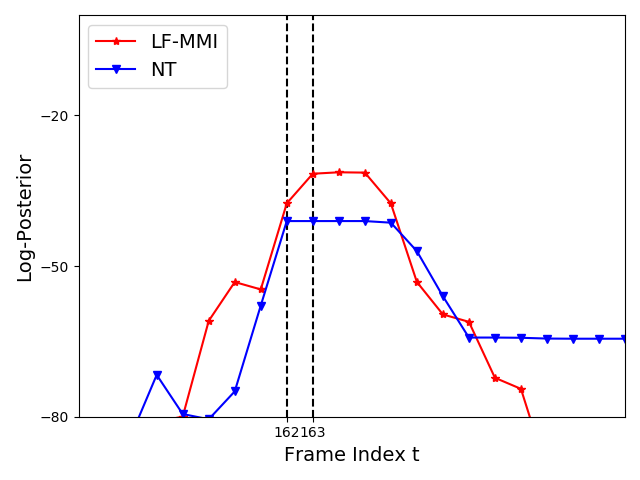}
    \caption{Log-posteriors $\log P(\textbf{W}_{1}^{u}|\textbf{O}_{1}^{t})$ provided by LF-MMI and transducer with various frame index $t$. $\textbf{W}_{1}^{u}$ is a correct partial hypothesis with $u=U-2$. (utterance BAC009S0745W0447 in dev set of Aishell-1).}
    \label{fig_mas_lookahead}
\end{figure}

Compared with the peak scores provided by the NT system, we observe that there is usually a time-dimensional delay in LF-MMI peak scores (see dashed lines in Fig.\ref{fig_mas_lookahead}). An explanation of this delay is: the LF-MMI criterion needs more frames to reach the peak score since more arcs for the newly proposed token are added in the numerator graph. By contrast, when a non-blank token is proposed by the NT system, its score can reach the peak value even without increasing the $t$ in the t-u alignment. 
To align the scores provided by NT and LF-MMI, a look-ahead mechanism is needed: it encourages the MMI criterion to provide an early validation against the newly proposed tokens during decoding. 

We also note, in each step when all proposed hypotheses are evaluated, scores of hypotheses that have identical $\textbf{W}_1^u$ but different alignment paths should be merged. But $S^{\tt{ali}}_{\tt{MMI}}$ should not participate in this process, since $S^{\tt{ali}}_{\tt{MMI}}$ directly assesses the validness of the aligned sequence pair $(\textbf{W}_1^u, \textbf{O}_1^t)$ and is the summed posterior of all alignment paths. 

We finally provide an overview of the NT beam search process with the MMI Alignment Score integrated into Algo. \ref{algo_nt}. The log-posteriors provided by the LF-MMI criterion are adopted whenever a $\textless blk\textgreater$ or a non-blank token is newly proposed. 
\rr{After the scores $s_t(\mathbf{W}_1^u)$ of each hypothesis is updated, the hypotheses with the identical blank-removed sequence but difference alignment paths should be merged\cite{alsd} using the function $CombineHypothesis(\cdot)$. Then, only $b$ hypotheses with highest scores can be kept in the $BeamPrune(\cdot)$ process before entering the next decoding step. After the decoding loop is ended, the hypotheses in the finished hypothesis set are sorted by their scores and returned.}

\begin{algorithm}[htpb]
\caption{NT beam search with MMI Alignment Score}
\begin{algorithmic}[1]
\label{algo_nt}
\REQUIRE acoustic feature sequence $\mathbf{O}$, vocabulary $\mathbf{V}$
\REQUIRE beam size $b$, maximum hypothesis length $U_{max}$
\REQUIRE decoding weight $\beta_{\tt{MMI}}$
\STATE Initial Hypothesis Set $\Omega_1 \gets \{(<sos>, 0)\}$
\STATE Finished  Hypothesis Set $\Omega_F \gets \emptyset$ 
\STATE Decoding Step $l \gets 1$
\FOR {$l = 1, ..., T+U_{max}$}
\STATE $\Omega_{l+1} \gets \emptyset$
\FOR {$(\mathbf{W}_1^{u}, s_t(\mathbf{W}_1^{u})) \in \Omega_{l}$}

\IF {$t >T$}
\STATE continue
\ENDIF

\STATE $s_{t+1}(\mathbf{W}_1^{u}) \gets s_t(\mathbf{W}_1^{u}) + \log p^{\tt{NT}}_{t+1}(\textless blk\textgreater|\mathbf{W_1^{u}}, \mathbf{O})$
\STATE $s_{mmi} \gets S^{\tt{ali}}_{\tt{MMI}}(\textbf{W}_{1}^{u},\textbf{O}_{1}^{t+1}) - S^{\tt{ali}}_{\tt{MMI}}(\textbf{W}_{1}^{u},\textbf{O}_{1}^{t})$
\STATE $s_{t+1}(\mathbf{W}_1^{u}) \gets s_{t+1}(\mathbf{W}_1^{u}) + \beta_{\tt{MMI}} \cdot s_{mmi}$

\STATE $\Omega_{l+1} \gets \Omega_{l+1} \cup (\mathbf{W}_1^u, s_{t+1}(\mathbf{W}_1^{u}))$ 

\IF {$t==T$}
\STATE $\Omega_{F} \gets \Omega_{F} \cup (\mathbf{W}_1^u, s_{t+1}(\mathbf{W}_1^{u}))$
\ENDIF

\FOR {$w_{u+1} \in \mathbf{W}$}
\STATE $\mathbf{W}_1^{u+1} \gets \mathbf{W}_1^{u} + w_{u+1}$
\STATE $s_{t}(\mathbf{W}_1^{u+1}) \gets s_{t}(\mathbf{W}_1^{u}) + \log p^{\tt{NT}}_{t}(w_{u+1}|\mathbf{W_1^{u}}, \mathbf{O})$
\STATE $s_{mmi} \gets S^{\tt{ali}}_{\tt{MMI}}(\textbf{W}_{1}^{u+1},\textbf{O}_{1}^{t}) - S^{\tt{ali}}_{\tt{MMI}}(\textbf{W}_{1}^{u},\textbf{O}_{1}^{t})$
\STATE $s_{t}(\mathbf{W}_1^{u+1}) \gets s_{t}(\mathbf{W}_1^{u+1}) + \beta_{\tt{MMI}} \cdot s_{mmi}$
\STATE $\Omega_{l+1} \gets \Omega_{l+1} \cup (\mathbf{W}_1^{u+1}, s_{t}(\mathbf{W}_1^{u+1}))$
\ENDFOR

\ENDFOR
\STATE CombineHypothesis($\Omega_{l+1}$)
\STATE BeamPrune($\Omega_{l+1}, b$)
\ENDFOR
\RETURN Sorted$(\Omega_F)$
\end{algorithmic}
\end{algorithm}

\subsection{LF-MMI Rescoring}
\label{rescore}
We further propose a unified rescoring method called LF-MMI Rescoring for both AEDs and NTs that are jointly optimized with the LF-MMI criterion.
Compared with using $S^{\tt{pref}}_{\tt{MMI}}$ and $S^{\tt{ali}}_{\tt{MMI}}$ in beam search, rescoring method is more computationally efficient. 

\begin{figure}
    \centering
    \includegraphics[width=8cm]{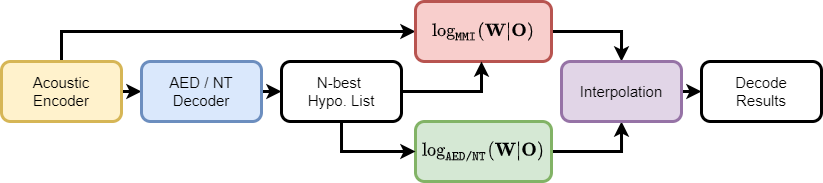}
    \caption{Diagram for MMI Rescoring. Interpolation from original posteriors and MMI posteriors are used for the final decision.}
    \label{fig_rescore}
\end{figure}

Assume the AED or NT system has been optimized by the LF-MMI criterion before decoding. 
As illustrated in Fig.\ref{fig_rescore}, the N-best hypothesis list is firstly generated by beam search without the LF-MMI criterion. 
Along this process, the log-posterior of each hypothesis $\textbf{W}$, namely $\log P_{\tt{AED/NT}}(\mathbf{W}|\mathbf{O})$, is also calculated. 
Next, another log-posterior for each hypothesis in the N-best hypothesis list, $\log P_{\tt{LF-MMI}}(\mathbf{W}|\mathbf{O})$, is computed according to the LF-MMI criterion. 
Finally, the interpolation of the two log-posteriors are calculated as follows:
\begin{equation}
\label{eq_rescore}
\begin{aligned}
    \log P(\mathbf{W}|\mathbf{O}) =& \log P_{\tt{AED/NT}}(\mathbf{W}|\mathbf{O}) \\
    &+ \lambda_{\tt{MMI}}  \cdot \log P_{\tt{LF-MMI}}(\mathbf{W}|\mathbf{O})
\end{aligned}
\end{equation}

where $\lambda_{\tt{MMI}}$ is the interpolation weight used in LF-MMI Rescoring.
As the LF-MMI criterion is applied to the acoustic encoder, MMI Rescoring can better emphasize the validness of hypotheses from the perspective of acoustics.
Moreover, since the denominator score $P(\mathbf{O}|\mathbb{G}_{den})$ is independent of the hypotheses, it can be considered as a constant for different hypotheses of a given utterance. 
Thus, only the numerator scores need to be calculated during LF-MMI Rescoring:
\begin{equation}
\setlength\abovedisplayskip{0cm}
\setlength\belowdisplayskip{-0.5cm}
    \log P_{\tt{MMI}}(\mathbf{W}|\mathbf{O}) = \log P(\mathbf{O}|\mathbb{G}_{num}(\mathbf{W})) - constant
\end{equation}

\subsection{Compare with MBR-based methods}
\label{compare_mbr}
Here the proposed LF-MMI training and decoding method is compared with its MBR counterparts. 
We claim that the three deficiencies of MBR-based methods discussed in section \ref{mbr_def} are solved or eschewed in our method.

\subsubsection{Consistent training and decoding}
The LF-MMI method achieves the consistency between training and decoding since the LF-MMI criterion is applied in both stages. By contrast, the MBR methods only consider the training process and apply no discriminative criterion during decoding.

\subsubsection{Train from randomly initialized model}
To compute the Bayesian risk, the MBR-based methods require a pre-trained model for on-the-fly decoding. However, this is not necessary for the LF-MMI training method since the on-the-fly decoding process is no longer needed. Instead, the LF-MMI criterion, along with the non-discriminative criteria, is capable to train from a randomly initialized model. 

\subsubsection{Training efficiency}
The MBR-based methods are slow in training speed due to the adoption of the sampling process (usually the on-the-fly decoding process) over the hypothesis space. The LF-MMI training method, however, replaces the sampling process with the \textit{forward-backward} algorithm on FSAs, which can be effectively implemented by matrix operators and is much faster. 

\section{Experimental setup}
\label{exp_setup}
Experimental setup is described in this section. The datasets are described in section \ref{para_data}. The model, optimization and decoding configurations are described in section \ref{para_model}. The implementation of LF-MMI and MBR methods is presented in section \ref{para_mmi}.

\subsection{Datasets}
\label{para_data}
We evaluate the effectiveness of the proposed method on three Mandarin datasets: Aishell-1\cite{aishell1}, Aishell-2\cite{aishell2} and our internal dataset Oteam. All utterances with non-Mandarin tokens are excluded. To examine the proposed method on low-resource ASR tasks, we additionally split a subset of the Aishell-1 training corpus, called Aishell-s, in which only 70 utterances for each speaker are kept. The scales of these datasets range from 30 hours to 14.3k hours. Details of these datasets are listed in table \ref{tab_dataset}.

\begin{table}[htpb]
    \centering
        \begin{tabular}{|l|c|c|c|c|}
    \hline
    Dataset      & Train hrs. & Num. utts. & Vocabulary Size  \\
    \hline
    Aishell-s    & 30        & 23.8k      & 4231     \\ 
    Aishell-1    & 178       & 120k       & 4231     \\
    Aishell-2    & 1k        & 963k       & 5214      \\
    Oteam        & 14.3k     & 14.6M      & 6267      \\
    \hline
    \end{tabular}
    \vspace{5pt}
    \caption{Statistics of datasets used in experiments}
    \label{tab_dataset}
\end{table}

\subsection{Model, optimization and evaluation}
The proposed method is evaluated on both AED and NT systems. For all experiments, we adopt a unified model architecture, optimization and decoding settings as described below.
\label{para_model}

\subsubsection{AED}
For the AED system, a 12-layer Conformer \cite{conformer} encoder and a 6-layer transformer decoder are adopted. 
For all attention modules in encoder and decoder, the feed-forward dimension, the attention dimension and the number of attention heads are 2048, 256 and 4 respectively. 
The CNN module in each Conformer layer has 31 kernels.
All batch-normalization\cite{batch_norm} layers are replaced by group-normalization\cite{wu2018group} layers with group number of 2 for training efficiency. 
The acoustic input is 80-dim Fbank features plus 3-dim pitch features. The input features are down-sampled by a 2-layer CNN with the factor of 4 before being fed into the Conformer encoder.
The dimension of token embedding is 256.
This architecture consumes around 46M parameters in total.

\subsubsection{NT}
For the NT system, a 12-layer Conformer encoder, a single-layer LSTM prediction network and an MLP joint network are adopted. 
The encoder has the same architecture as that in AED except the down-sampling factor is 6. 
The dimension of token embedding and prediction network is 1024 and 512 respectively. The joint network is also in the size of 512. 
This architecture consumes around 90M parameters. 

\subsubsection{Optimization}
All experiments adopt the Adam optimizer\cite{adam} with warm-up steps of 25k and inverse square root decay schedule\cite{noam}. 
The global batch-size, peak learning rate, number of GPUs and number of epochs for different datasets are listed in table \ref{tab_bsz}. 
SpecAugment\cite{specaug} is consistently adopted with two time masks and two frequency masks.
We use the speed perturbation with factors 0.9, 1.0 and 1.1 for all Aishell datasets so their data scales are expanded by 3 times.
All models are trained on Tesla P40 GPUs.

\subsubsection{Evaluation}
We average the checkpoints from the last 10 epochs for evaluation. 
The decoding algorithms for AED and NT systems are presented in \cite{lasctc} and \cite{alsd} respectively. 
The beam size for all decoding stages is fixed to 10. During decoding, Word-level N-gram language models\cite{wngram} trained from the training transcriptions are optionally integrated with a default weight of 0.4.
Our implementation is mainly revised from Espnet\cite{espnet, espnet2020}.

\begin{table}[htpb]
    \centering
    \begin{tabular}{|l|c|c|c|c|}
    \hline
         Dataset     & Batch size & Peak LR & Num. GPUs & Num. epochs    \\
    \hline
         Aishell-s   & 64   & 3e-4  & 8 & 100 \\
         Aishell-1   & 64   & 3e-4  & 8 & 100 \\
         Aishell-2   & 512  & 3e-4  & 8 & 50  \\
         Oteam       & 1024 & 3e-3  & 32& 30  \\
    \hline
    \end{tabular}
    \vspace{5pt}
    \caption{Optimization settings for different datasets.}
    \label{tab_bsz}
\end{table}

\subsection{LF-MMI and MBR implementation}
\label{para_mmi}

\subsubsection{LF-MMI method implementation}
\rr{Our LF-MMI implementation mainly follows \cite{lfmmi_18}. 
In terms of topology, the CTC-topology\cite{ctc_topo} is adopted in all experiments. 
As discussed in section \ref{intro_lfmmi}, to reduce the memory and computational cost, phoneme is adopted as the modeling unit to compute the LF-MMI loss in this work, which means a phone-level lexicon is used. 
All lexicons of open-accessible datasets are from standard Kaldi recipes\footnote{https://github.com/kaldi-asr/kaldi/tree/master/egs/aishell}\footnote{https://github.com/kaldi-asr/kaldi/tree/master/egs/aishell2} and the lexicon in Aishell-2 is also used in Oteam dataset.
All phones are used in context-independent format. When compiling the lexicon into FST, optional silence with the probability of 0.5 is also added.
The order of the phone language model used in the FSA compilation is fixed to 2. No alignment information in any form is used.
The generation of numerator and denominator graphs still follows\cite{lfmmi_18} and the gradients are computed by standard forward-backward algorithm\cite{forwardbackward}.
}
By default, the training hyper-parameters $\alpha_{\tt{AED}}$ and $\alpha_{\tt{NT}}$ are empirically set to 0.3 and 0.5 respectively. 
The decoding hyper-parameters $\beta_{\tt{MMI}}$ and the rescoring paramter $\lambda_{\tt{MMI}}$ are set to 0.2 consistently. The \textit{look-ahead step $\tau$} in MMI Alignment Score computation is set to 3.
The LF-MMI criterion is based on differentiable FST algorithms\cite{diff_fst} and is implemented by k2\footnote{https://github.com/k2-fsa/k2}.
\subsubsection{MBR-based method implementation}
To implement the MBR training, we initialize our model from the last checkpoint of the baseline model (AED or NT models without LF-MMI). 
\rrr{The training lasts 10\% epochs described in table \ref{tab_bsz}, which is found necessary to achieve full convergence and obtain the benefit of model average.}
Due to the GPU memory limitation, we consistently adopt the beam size of 4 in all on-the-fly decoding processes. As MBR methods consume more GPU memory, we halve the mini-batch size and double the gradient accumulation number to ensure the same batch size. 
The smooth constant $\epsilon$ in Eq.\ref{mbr_appro} is set to 1e-10. 
Following the implementation of \cite{mbr_las, mbr_rnnt}, the non-discriminative training criteria of the AED and NT system are also adopted for regularization and are interpolated with the MBR criterion equally. 
Note we use the raw input features for on-the-fly decoding, which is not corrupted by SpecAugment.
As the number of epochs is greatly reduced in MBR training, we average the checkpoints from all epochs for evaluation. 

\section{Experimental Results and Analysis}
\label{exp_result}
This section is organized as follows: To begin with, an overview of the proposed method on two open-source datasets is provided in section \ref{exp_aishell} to show the strength of the proposed method against previous methods. To provide a detailed analysis of each component in the proposed method, the ablation study and the hyper-parameter investigation are conducted in section \ref{exp_ablation} and section \ref{exp_param} respectively. To verify the generalization of the proposed method on both small and large scale speech corpus, experiments on a 30-hour dataset and a 14.3k-hour dataset are then provided in section \ref{exp_tiny} and section \ref{exp_oteam}. Finally, to show the performance difference between the proposed LF-MMI method and previous MBR-based methods on E2E ASR, a comparison between these two methods is given in section \ref{exp_mbr} and section \ref{exp_mbr2}.

\subsection{Overall results on Aishell-1 and Aishell-2}
\label{exp_aishell}
In table \ref{tab_aishell_result}, we present our best results obtained by the proposed LF-MMI method on the open-sourced Aishell-1 and Aishell-2 datasets for both AED and NT systems. To show the strength of the proposed method, competitive results in the previous publication are also provided.

\begin{table}[htpb]
    \centering
    \begin{tabular}{|l|c|c|c|c|c|}
    \hline
         \multirow{2}{*}{System} &  \multicolumn{2}{|c|}{Aishell-1} & \multicolumn{3}{|c|}{Aishell-2}    \\
         \cline{2-6}
         & dev & test & android & ios & mic \\
         \hline\hline
         \multicolumn{6}{|l|}{Other Frameworks} \\
         \hline
         SpeechBrain\cite{speechbrain} & 5.60 & 6.04 & - & -  & -    \\
         Espnet\cite{espnet_result} & 4.4  & 4.7  & 7.6 & 6.8 & 7.4 \\
         Wenet\cite{wenet}          & -    & 4.36 & -   & 5.35& -   \\
         Icefall                  & -    & 4.26 & - & - & -  \\
         \hline\hline
         \multicolumn{6}{|l|}{Our Results} \\
         \hline
         AED (baseline)             & 4.33 & 4.71 & 6.07 & 5.29 & 5.88 \\
         \ \ + proposed method      & 4.08* & 4.45* & 5.92* & 5.15 & 5.77* \\
         NT  (baseline)             & 4.47 & 4.84 & 6.33 & 5.58 & 6.25 \\
         \ \ + proposed method      & \bf{3.79*} & \bf{4.10*} & \bf{5.85*} & \bf{5.02*} & \bf{5.66*} \\
         \hline
    \end{tabular}
    \vspace{5pt}
    \caption{Main results on Aishell-1 and Aishell-2 datasets. CER is reported. All our results are with word language models. * means statistically significant improvement compared with baseline systems.}
    \label{tab_main_result}
\end{table}

As suggested in table \ref{tab_main_result}, the proposed LF-MMI training and decoding methods achieve significant performance improvement on both AED and NT systems and on all evaluation sets. For the NT system, up to 15.3\% relative error reduction is achieved on the Aishell-1 test set and the absolute CER reductions on all evaluation sets are above 0.48\%. Also, up to 5.8\% relative CER reduction (Aishell-1 dev set) is also achieved on the AED system. 
The significance of these improvements is further confirmed by matched pairs sentence-segment word error (MAPSSWE) based significant test\footnote{https://github.com/talhanai/wer-sigtest} between the baseline systems and the systems with our LF-MMI method. Given the significant level of $p=0.01$, the proposed method achieve statistically significant improvement on most evaluation sets.
Furthermore, to the best of our knowledge, the NT system with the proposed method outperforms all other public-known results and achieves state-of-the-art (SOTA) CERs on both Aishell-1 and Aishell-2 datasets.

\subsection{Ablation Study}
\begin{table*}[htpb]
    \centering
    \begin{tabular}{|c|l|c|c|c|c|c|c|c|c|c|c|}
    \hline
         \multirow{3}{*}{No.} & \multirow{3}{*}{System} &  \multicolumn{4}{|c|}{Aishell-1 (178 hrs)} & \multicolumn{6}{|c|}{Aishell-2 (1k hrs)}  \\
         \cline{3-12}
         & & \multicolumn{2}{|c|}{w/o LM} & \multicolumn{2}{|c|}{w LM} & \multicolumn{3}{|c|}{w/o LM} & \multicolumn{3}{|c|}{w LM} \\
         \cline{3-12}
         && dev & test & dev & test & android & ios & mic & android & ios & mic \\
         \hline
         \multicolumn{12}{|l|}{Attention-Based Encoder-Decoder (AED)} \\
         \hline
         1 & Attention + char. CTC (baseline)    & 4.60      & 5.07      & 4.33      & 4.71      & \bf{6.60}      & \bf{5.72} & 6.58      & 6.07      & 5.29      & 5.88 \\
         2 & Attention + LF-MMI                  & -         & -         & -         & -         & -         & -         & -         & -         & -         & -    \\
         3 & \ \ + MMI Prefix Score              & 4.83      & 5.36      & 4.32      & 4.83      & 7.42      & 6.08      & 6.91      & 6.58      & 5.47      & 6.17 \\
         4 & Attention + char. CTC + phone CTC   & 4.69      & 5.24      & 4.40      & 4.85      & 6.77      & 5.83      & 6.67      & 6.27      & 5.49      & 6.05 \\
         5 & Attention + char. CTC + LF-MMI      & \bf{4.55} & \bf{5.05} & 4.22      & 4.59      & 6.69 & 5.80      & \bf{6.44} & 5.98      & 5.28      & 5.83 \\
         6 & \ \ + MMI Prefix Score              & \bf{4.55} & 5.10      & \bf{4.08} & \bf{4.45} & 6.75      & 5.80      & 6.46      & \bf{5.88} & \bf{5.15} & \bf{5.77} \\ 
         7 & \ \ + LF-MMI Rescoring              & \bf{4.55} & 5.10      & \bf{4.08} & 4.49      & 6.75      & 5.80      & 6.46      & 5.92      & \bf{5.15} & \bf{5.77} \\
         \hline
         \multicolumn{12}{|l|}{Neural Transducer (NT)} \\
         \hline
         8 & Transducer (baseline)               & 4.41 & 4.82 & 4.47 & 4.84 & 6.52 & 5.81 & 6.52 & 6.33 & 5.58 & 6.25 \\
         9 & Transducer + Phone CTC              & 4.81 & 5.24 & 4.59 & 4.96 & 7.08 & 6.01 & 6.86 & 6.94 & 6.00 & 6.75 \\
         10& Transducer + LF-MMI                 & 4.37 & 4.86 & 4.27 & 4.73 & 6.64 & 5.80 & 6.57 & 6.33 & 5.52 & 6.25 \\
         11& \ \ + MMI Alignment Score           & 4.29 & 4.81 & 3.92 & 4.38 & 6.58 & 5.70 & 6.56 & 5.97 & 5.17 & 5.88 \\
         12& \ \ + LF-MMI Rescoring              & 4.28      & 4.78      & 3.98      & 4.46      & 6.58      & 5.73      & 6.57      & 6.05 & 5.24 & 5.94 \\
         13& Transducer + char. CTC              & 4.91      & 5.02      & 4.88      & 4.88      & 6.45      & 5.47      & 6.26      & 6.23 & 5.25 & 6.04 \\
         14& Transducer + char. CTC + phone CTC  & 4.54      & 5.04      & 4.39      & 4.82      &  6.62 &5.58 & 6.45 & 6.49 & 5.47 & 6.23 \\
         15& Transducer + char. CTC + LF-MMI     & 4.25      & 4.64      & 4.08      & 4.37      & 6.47      & 5.51      & 6.29      & 6.14 & 5.28 & 5.97 \\
         16& \ \ + MMI Alignment Score           & 4.18      & \bf{4.54} & \bf{3.79} & \bf{4.10} & \bf{6.41} & \bf{5.40} & 6.18      & \bf{5.85} & \bf{5.02} & \bf{5.66} \\
         17& \ \ + LF-MMI Rescoring              & \bf{4.17} & 4.57      & 3.84      & 4.21      & \bf{6.41} & 5.42      & \bf{6.17} & 5.95 & 5.07 & 5.70 \\
         \hline
    \end{tabular}
    \vspace{5pt}
    \caption{Detailed results on AISHELL-1 and AISHELL-2 datasets. CER is reported. Word N-gram LMs are optionally used.}
    \label{tab_aishell_result}
\end{table*}
\label{exp_ablation}
To systematically analyze the effectiveness of each component in the proposed LF-MMI training and decoding method, an ablation study is conducted from various perspectives on Aishell-1 and Aishell-2 datasets. The experimental results are reported in table \ref{tab_aishell_result}. Our main observations are listed as follows.

\subsubsection{LF-MMI training method and CTC regularization}
For all experiments conducted on Aishell-1 and Aishell-2, token-level CTC criterion is found as a necessary regularization during training.
As suggested in table \ref{tab_aishell_result}, if the CTC regularization is not adopted, the impact of using LF-MMI as an auxiliary criterion is marginal on the NT system (exp.8 vs. exp.10) and is even harmful to the AED system (exp.1 vs. exp.3). 
However, performance improvement can be obtained on most evaluation sets (exp.5, exp.15) when the CTC regularization is adopted. 
Typically, the adoption of LF-MMI criterion with CTC regularization provides much considerable CER reduction on the NT system. E.g., in exp.15, CER on Aishell-1 test set is reduced from 4.84\% to 4.47\%.

One possible explanation for the CTC regularization is that the LF-MMI criterion is sensitive to over-fitting and requires regularization to alleviate this problem. 
In previous literature working on DNN-HMM systems \cite{lfmmi_16,lfmmi_18}, the LF-MMI criterion is also found susceptible to over-fitting and the frame-level cross-entropy (CE) criterion is adopted as the indispensable regularization.
Similarly, the token-level CTC criterion can be considered as another regularization for LF-MMI in E2E frameworks.

\subsubsection{LF-MMI decoding methods}
The results in table \ref{tab_aishell_result} also suggest that consistently using the LF-MMI criterion in both training and decoding stages can further improve the recognition performance (exp.5 vs. exp.6,7; exp.10 vs.exp.11,12; exp.15 vs. exp.16,17). In addition, the on-the-fly decoding methods slightly outperform the rescoring methods (exp.6 vs.exp.7; exp.11 vs. exp.12; exp.16 vs. exp.17).

\subsubsection{Impact of phone-level information}
As our implementation of the LF-MMI criterion still adopts phone-level information (the external lexicon), one may challenge that the improvement of LF-MMI training should be attributed to this additional information. To address this concern, the LF-MMI criterion is replaced by the phone-level CTC criterion with the same lexicon for comparison. Compared with the baseline systems, the adoption of phone-level CTC criterion during training provides no performance improvement (exp.1 vs. exp.4; exp.8 vs. exp.9,14), which indicates that the potential benefit of additional phone-level information is limited and the improvement achieved by LF-MMI is from its discriminative nature.

\subsubsection{Impact of language models}
Word N-gram LMs are optionally adopted in all experiments and achieve consistent performance improvement. 
However, the improvement provided by the LMs varies for different E2E ASR frameworks. 
In AED systems, significant CER reduction is achieved by word N-gram LMs no matter LF-MMI criterion is adopted or not. 
On the other hand, the adoption of word N-gram LMs provides larger performance improvement on NT system trained by the LF-MMI criterion.

\subsection{Impact of hyper-parameters}
\begin{table*}[htpb]
    \centering
    \begin{tabular}{|c|l|c|c|c|c|c|c|c|}
    \hline
    \multirow{2}{*}{No.} & \multirow{2}{*}{System} & Train & Decoding & look-ahead &\multicolumn{2}{|c|}{Beam Search} & \multicolumn{2}{|c|}{LF-MMI Rescoring} \\
    \cline{6-9}
    & & Weight & Weight & step & dev & test & dev & test\\ 
    \hline
    1 & Attention. + CTC (AED, baseline) & - & - & - & 4.33 & 4.71 & 4.33 & 4.71\\
    \cline{2-9}
    2 & \multirow{9}{*}{AED + LF-MMI}              & 0.1 & \multirow{4}{*}{0.2} &  \multirow{4}{*}{-} & 4.11 & 4.57 & 4.13 & 4.60\\
    3 &                                            & 0.3 &                      && \bf{4.08} & \bf{4.45} & \bf{4.08} & \bf{4.49} \\
    4 &                                            & 0.5 &                      && 4.09 & 4.51 & 4.10 & 4.54 \\
    5 &                                            & 0.8 &                      && 4.11 & 4.56 & 4.12 & 4.58 \\
    \cline{3-9}
    6 &                                            & \multirow{5}{*}{0.3} & 0.0 & \multirow{5}{*}{-} & 4.22 & 4.59 & 4.22 & 4.59\\
    7 &                                            &                      & 0.05&& 4.16 & 4.51 & 4.17 & 4.53\\
    8 &                                            &                      & 0.1 && 4.13 & 4.49 & 4.13 & 4.51\\
    9 &                                            &                      & 0.2 && 4.08 & \bf{4.45} & 4.08 & 4.49\\
    10&                                            &                      & 0.3 && \bf{4.06} & 4.46 & \bf{4.07} & \bf{4.48}\\
    \hline\hline
    11& Transducer (NT, baseline)                            & -   & - & 0                & 4.20 & 4.60 & 4.20 & 4.60\\
    \cline{2-9}
    12& \multirow{10}{*}{NT + LF-MMI}               & 0.1 & \multirow{4}{*}{0.2} & \multirow{4}{*}{3} & 3.82 & 4.23 & 3.91 & 4.36\\
    13&                                            & 0.3 &                      && \bf{3.77} & 4.18 & \bf{3.86} & 4.29 \\
    14&                                            & 0.5 &                      && 3.79 & \bf{4.10} & \bf{3.86} & \bf{4.21}\\
    15&                                            & 0.8 &                      && 4.01 & 4.22 & 3.90 & 4.27\\
    \cline{3-9}
    16&                                            & \multirow{6}{*}{0.5}       & 0.0 & \multirow{5}{*}{3} & 4.08 & 4.37    & 4.08 & 4.37 \\
    17&                                            &                            & 0.05&& 3.93 & 4.24 & 4.04 & 4.35 \\
    18&                                            &                            & 0.1 && 3.85 & 4.17 & 3.93 & 4.27 \\
    19&                                            &                            & 0.2 && \bf{3.79} & \bf{4.10} & 3.86 & 4.21 \\
    20&                                            &                            & 0.3 && 3.86 & 4.14 & \bf{3.84} & \bf{4.17} \\
    \cline{4-9}
    21 &                                           &                            & 0.2 & 0 & 3.82 & 4.14 & - & - \\ 
    \hline
    \end{tabular}
    \vspace{5pt}
    \caption{CER\% results on Aishell-1 with different training weights ($\alpha_{\tt{AED}}, \alpha_{\tt{NT}}$), decoding weights ($\beta_{\tt{MMI}}, \gamma_{\tt{MMI}}$) and look-ahead steps $\tau$ of the proposed LF-MMI method. All results are with word N-gram LMs.}
    \label{tab_param}
\end{table*}

\label{exp_param}

In this part, we investigate the impact of the training and decoding hyper-parameters on the Aishell-1 dataset and present the results in table \ref{tab_param}. 
Based on these results, we claim the proposed method maintains superior robustness against hyper-parameters. During the training stage of both AED and NT systems, adopting LF-MMI as an auxiliary criterion is consistently helpful with various weights in \{0.1, 0.3, 0.5, 0.8\} (exp.1 vs. exp.2-5; exp.11 vs. exp.12-15). Additionally, the adoption of LF-MMI decoding methods, including both the beam search methods and the rescoring method, also provides consistent CER reduction with all weights in \{0.05, 0.1, 0.2, 0.3\} (exp.6 vs. exp.7-10; exp.16 vs. exp.17-20). These results suggest that the proposed LF-MMI training and decoding methods may lead to improvement without much engineering effort.

Also note the training weights of 0.3 (for AED), 0.5 (for NT) and decoding weights of 0.2 (for both AED and NT) achieve much promising results, which are the default settings in the experiments aforementioned. Finally, exp.21 suggests that the adoption of the look-ahead mechanism in the MMI Alignment Score is beneficial, even though the improvement is marginal.

\subsection{Results on the low-resource dataset}
\label{exp_tiny}
To further show the generalization ability of the LF-MMI-based methods on the low-resource dataset, experiments conducted on the 30-hour Aishell-s dataset are shown in table \ref{tab_aishell_s}. Note that the original dev and test sets of Aishell-1 are still used for evaluation in this experiment.

\begin{table}[b]
    \centering
    \begin{tabular}{|l|c|c|c|c|}
    \hline
          \multirow{3}{*}{System} &  \multicolumn{4}{|c|}{Aishell-s (30 hrs)} \\
         \cline{2-5}
         & \multicolumn{2}{|c|}{w/o LM} & \multicolumn{2}{|c|}{w LM} \\
         \cline{2-5}
         & dev & test & dev & test \\
         \hline
         \multicolumn{5}{|l|}{Attention-Based Encoder-Decoder (AED)} \\
         \hline
          Attention + char. CTC (baseline)    & 9.70 & 10.28   & 9.16 & 9.59 \\
          Attention + char. CTC + LF-MMI      & 9.17 & 9.94    & 8.37 & 8.94 \\
          \ \ + MMI Prefix Score              & \bf{9.01} & \bf{9.87}    & \bf{7.80} & \bf{8.47} \\
          \ \ + LF-MMI Rescoring              & 9.02 & \bf{9.87}    & 7.92 & 8.57 \\
         \hline
         \multicolumn{5}{|l|}{Neural Transducer (NT)} \\
         \hline
          Transducer (baseline)                & 11.37 & 12.22 & 11.09 & 11.86 \\
          Transducer + char. CTC + LF-MMI      & 10.17 & 11.05 & 9.87  & 10.59 \\
          \ \ + MMI Alignment Score            & \bf{9.40}  & 10.32 & \bf{8.82}  & \bf{9.53}\\
          \ \ + LF-MMI Rescoring               & 9.45  & \bf{10.28} & 9.02  & 9.73 \\
         \hline
    \end{tabular}
    \vspace{5pt}
    \caption{CER\% results of the models trained by 20\% Aishell-1 training data. Word N-gram LMs are optionally used.}
    \label{tab_aishell_s}
\end{table}

Consistent with the observations in section \ref{exp_ablation}, we find that the proposed LF-MMI method generalizes well on the low-resource scenario. As shown in table \ref{tab_aishell_s}, for the NT system, the training and decoding methods provide absolute CER reductions of 1.27\% and 1.06\% respectively and the total relative improvement is 19.6\%. For the AED system, the relative improvement achieved by the proposed method is 11.7\%. Interestingly, the relative improvement of the proposed LF-MMI method is even larger on the low-resource data.

\subsection{Results on the large-scale dataset}
\begin{table*}[t]
    \centering
    \begin{tabular}{|c|l|c|c|c|c|c|c|c|c|c|c|c|c|}
    \hline
         \multirow{2}{*}{No.} & \multirow{2}{*}{System} & \multicolumn{6}{|c|}{w/o LM} & \multicolumn{6}{|c|}{w LM} \\
         \cline{3-14}
         && st & tv & mu & ed & re & Mean & st & tv & mu & ed & re & Mean\\
         \hline 
         \multicolumn{14}{|l|}{Attention-Based Encoder-Decoder (AED)} \\
         \hline
         1 & Attention + CTC (baseline)      & 17.20 & 9.24 & 17.24 & 11.43 & 5.44 & 12.11 &  16.30 & 7.71 & 14.66 & 11.42 & 4.98 & 11.01 \\
         2 & Attention + LF-MMI              & - & - & - & - & - & - & - & - & - & - & - & - \\
         3 & \ \ + MMI Prefix Score          & \bf{16.85} & \bf{8.32} & \bf{14.75} & \bf{10.87} & \bf{4.98} & \bf{11.15} &  \bf{16.00} & \bf{7.10} & 14.16 & \bf{11.02} & \bf{4.88} & 10.63 \\
         4 & Attention + CTC + LF-MMI        & 17.13 & 8.69 & 15.48 & 11.26 & 5.38 & 11.59 &  16.14 & 7.38 & \bf{13.25} & 11.28 & 4.92 & \bf{10.59} \\
         5 & \ \ + MMI Prefix Score Decoding & 17.20 & 8.72 & 15.51 & 11.25 & 5.37 & 11.61 &  16.20 & 7.44 & 13.32 & 11.15 & 4.94 & 10.61 \\
         6 & \ \ + LF-MMI Rescoring          & 17.16 & 8.73 & 15.46 & 11.27 & 5.34 & 11.59 &  16.17 & 7.43 & 13.29 & 11.28 & 4.91 & 10.62 \\
         \hline
         \multicolumn{10}{|l|}{Neural Transducer (NT)} \\
         \hline
         7 & Transducer (baseline)                & 17.00 & 8.28 & 15.23 & 11.43 & 5.00 & 11.39 & 16.08 & 7.10 & 12.83 & 11.09 & 4.66 & 10.35 \\
         8 & Transducer + LF-MMI                  & 16.68 & 7.51 & 15.18 & 10.86 & 4.67 & 11.00 & 15.89 & 6.27 & 12.83 & 10.79 & 4.40 & 10.04 \\
         9 & \ \ + MMI Alignment Score Decoding   & 16.50 & 7.42 & 14.94 & \bf{10.78} & 4.66 & 10.86 & 15.88 & 6.26 & 12.58 & \bf{10.70} & 4.39 & 9.96  \\
         10 & \ \ + MMI Rescoring                 & 16.51 & \bf{7.35} & 13.93 & 10.83 & \bf{4.63} & 10.65 & \bf{15.61} & \bf{6.19} & 11.38 & 11.21 & \bf{4.27} & \bf{9.73}  \\
         11& Transducer + char. CTC + LF-MMI      & 16.54 & 7.42 & 14.45 & 11.23 & 4.81 & 10.89 & 15.74 & 6.60 & 12.18 & 11.29 & 4.58 & 10.08 \\
         12& \ \ + MMI Alignment Score Decoding   & 16.57 & 7.42 & \bf{11.36} & 11.12 & 4.82 & \bf{10.26} & 15.75 & 6.60 & 12.23 & 11.23 & 4.56 & 10.07 \\
         13& \ \ + LF-MMI Rescoring               & \bf{16.40} & 7.42 & 13.59 & 11.24 & 4.80 & 10.69 & 15.50 & 6.60 & \bf{11.05} & 11.56 & 4.44 & 9.83  \\ 
         \hline
    \end{tabular}
    \vspace{5pt}
    \caption{Results on 14.3k-hour Oteam dataset. CERs of 5 test sets are reported. Word N-gram LMs are optionally used.}
    \label{tab_oteam_result}
\end{table*}

\label{exp_oteam}
To further show the effectiveness of the proposed method on the large-scale dataset, experiment conducted on the internal 14.3k-hour Oteam dataset is further presented in this section.
In this experiment, five test sets are provided to assess the model performance in various real applications: speech translation (st), television (tv), music (mu), education (ed) and reading (re). We change the decoding coefficient $\beta_{\tt{MMI}}$ and rescoring coefficient $\gamma_{\tt{MMI}}$ to 0.05 while keeping the other hyper-parameters unchanged. All results are presented in table \ref{tab_oteam_result}. Our main observations on this dataset are reported as follows.

\subsubsection{Overall improvement}
The proposed LF-MMI training and decoding method is still beneficial in large-scale scenario. For the AED system, our method provides an absolute CER reduction of 7.9\% on average (exp.1 vs. exp.3) while this number for the NT system is 9.9\% (exp.7 vs. exp.12). The maximum improvement is observed on the music (mu) test set, where the CER is reduced from 15.23\% to 11.36\%. 

\subsubsection{Impact of CTC regularization}
In the large-scale experiment, it is no longer that crucial to adopt CTC regularization during training. For the AED system, the model performance is consistently compromised by the adoption of CTC regularization (exp.3 vs. exp.4-6).  For the NT system, the averaged CERs achieved by the systems with or without CTC regularization show limited difference. We suppose the LF-MMI criterion can work independently when the data scale is comparatively large as the over-fitting problem is alleviated.

\subsection{Compare with MBR-based Method} 
\label{exp_mbr}
\begin{table}[htpb]
    \centering
    \begin{tabular}{|l|c|c|c|c|c|c|c|c|}
         \hline
          \multirow{2}{*}{System} & \multicolumn{2}{|c|}{Aishell-1} & \multicolumn{3}{|c|}{Aishell-2} & relative\\
         \cline{2-6}
           & dev & test & android & ios & mic & train \rr{time} \\
         \hline
          AED          & 4.33 & 4.71 & 6.07 & 5.29 & 5.88 & 1.00 \\
          \ + MBR     & 4.30 & 4.70 & 6.36 & 5.33 & 6.02  & \rr{5.81} \\
          \ + LF-MMI   & \bf{4.08} & \bf{4.45} & \bf{5.92} & \bf{5.15} & \bf{5.77} & \rr{1.62}  \\
         \hline
          NT           & 4.47 & 4.84 & 6.33 & 5.58 & 6.25 & 1.00\\
          \ + MBR      & 4.48 & 4.88 & 6.52 & 5.72 & 6.37 & \rr{2.78}\\
          \ + LF-MMI    & \bf{3.79} & \bf{4.10} & \bf{5.85} & \bf{5.02} & \bf{5.66} & \rr{1.77} \\
         \hline 
    \end{tabular}
    \vspace{5pt}
    \caption{Comparison between the proposed MMI method and MBR methods. Relative training speed on \rr{Aishell-2} dataset and CERs on AISHELL-1 and Aishell-2 datasets are reported.}
    \label{tab_mbr}
\end{table}
To compare the proposed LF-MMI method with the MBR-based methods, we conduct MBR training methods on both AED and NT systems over Aishell-1 and Aishell-2 datasets. All results are presented in table \ref{tab_mbr}. 

In these experiments, the proposed LF-MMI method outperforms the MBR-based methods on both accuracy and speed consistently. Compared with the baseline systems, the MBR-based methods achieve marginal improvement on Aishell-1 AED experiment and encounter degradation in all other experiments. By contrast, the proposed LF-MMI method achieves consistent and considerable performance improvement on both datasets and both frameworks. 

\rr{Besides the recognition performance, the total training time (including the first-stage non-discriminative training and the second-stage discriminative training) of the MBR-based methods is also compared with that of the proposed LF-MMI method. As suggested in the table, the MBR-based methods are slower than the proposed LF-MMI methods.} 

\subsection{Further Analysis on MBR-Based Methods} 
\label{exp_mbr2}
\begin{figure*}
    \centering
    \includegraphics[width=\linewidth]{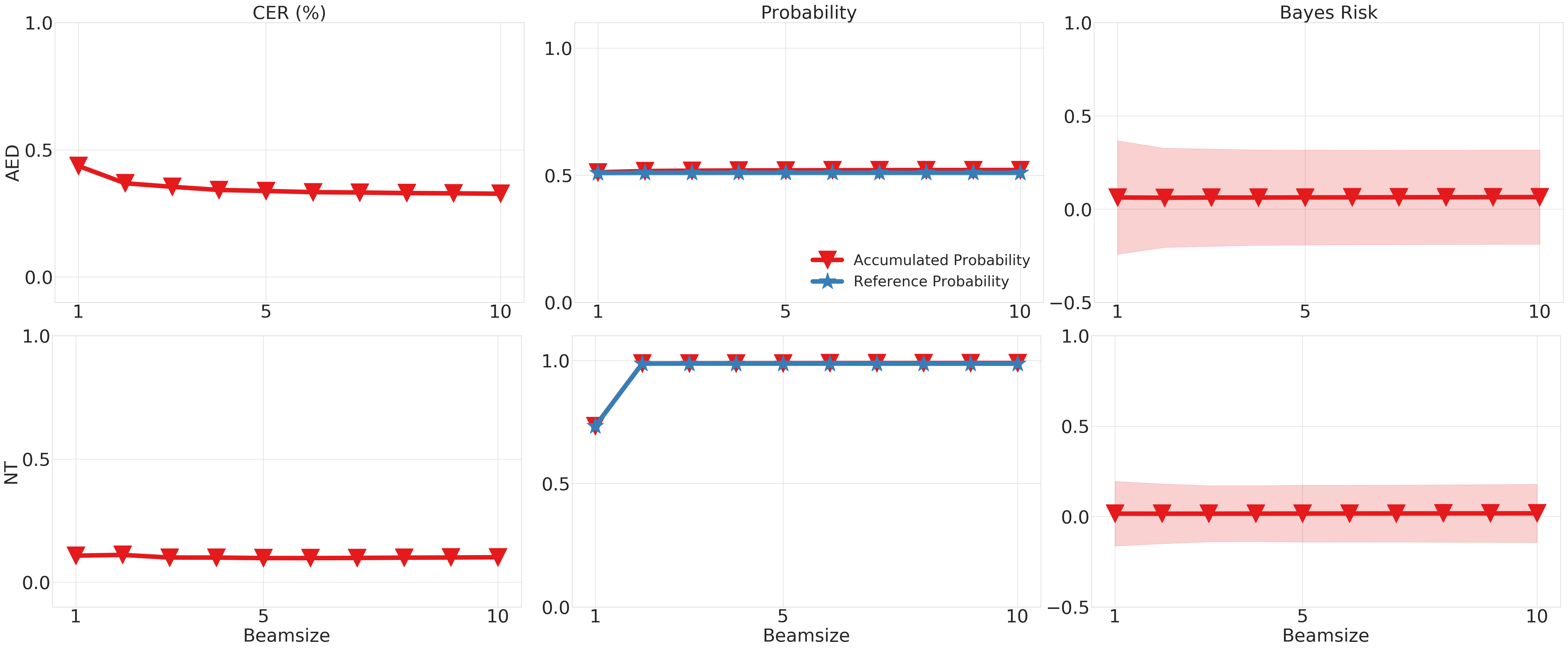}
    \caption{Statistics of the on-the-fly decoding results over the training sets of Aishell-1 dataset. The upper row is for the AED system and the downer row is for NT systems. Left column : CER\%. Middle column: the mean value of the accumulated probability over the N-best hypothesis list and the probability of the reference token sequence. Right column: the mean value of the approximated Bayesian risk and its standard deviation.}
    \label{fig_mbr_stat}
\end{figure*}

Although improvements are consistently observed in previous literature \cite{mbr_las, mwer_las, mwer_rnnt, mbr_rnnt}, we can hardly achieve better CER results on either AED or NT systems with these MBR-based methods. 
To find the reasons, further investigation is conducted on the MBR-trained models with Aishell-1 dataset. Specifically, the results of on-the-fly decoding over the training set are analyzed and the statistics are presented in Fig. \ref{fig_mbr_stat}. Our main observations are reported as below.

\subsubsection{CER over the training set}
As shown in the left column of Fig. \ref{fig_mbr_stat}, for all models trained by MBR criterion, the CER over the training set is very small (all below 0.5\%). This small CER suggests the MBR-trained models have fit the training data to a satisfactory level. Note the motivation of MBR-based methods is to reduce the expected CER, this observation also verifies the MBR-based methods have fulfilled their original intention and reduced the CER over the training set effectively.

\subsubsection{Probability of reference token sequence}
As presented in the middle column of Fig. \ref{fig_mbr_stat}, the accumulated probability of the N-best hypothesis list and the probability of the reference token sequence are much close, which means the probability occupied by the reference token sequence is dominant. This observation suggests the posterior distribution output by the MBR-trained models is very sharp and these MBR-trained models are much confident in the reference token sequence.

\subsubsection{Approximated Bayesian risk}
With the two observations above, the approximated Bayesian risk in Eq.\ref{mbr_appro} tends to be very small, as the accumulated probability in the denominator is considerably large but the risk of the erroneous hypotheses with small probability is marginal in the numerator. This is also experimentally verified in the right column of Fig. \ref{fig_mbr_stat}: the mean value of the approximated Bayesian risk is less than 0.07 for the AED system and 0.02 for the NT system.

Given these observations, the MBR-trained models are believed to have fit the training data successfully: the approximated Bayesian risk objective becomes very small after this training. In addition, the adoption of larger beam size during the on-the-fly decoding provides limited increment on the Bayesian risk objective.
However, given this promising results over the training set, generalizing these MBR-trained models to the evaluation sets still achieves limited improvement. This ineffectiveness may be attributed to the over-fitting problem: take the NT system as an example, the CER over test set is more than 10 times larger than that of the training set. By contrast, we observe this ratio for LF-MMI objective in both AED and NT systems is below 3.

\section{Discussion and conclusion}
\label{conclusion}
This work proposes to integrate the discriminative training criterion, LF-MMI, into two of the most popular E2E ASR systems: AED and NT. 
Unlike the previous MBR-based methods that only consider the training stage, the proposed method can be consistently applied to training and decoding for better performance. 
Specifically, the LF-MMI criterion acts as an auxiliary criterion during training. 
In decoding, two algorithms are proposed for the on-the-fly decoding process of AED and NT systems respectively and a unified rescoring method using LF-MMI criterion is also provided. 

In experiments, we show that the proposed methods lead to consistent improvement. 
The training and decoding methods are carefully examined on 4 datasets ranging from tens hours to more than 10k hours.
The impact of regularization, language model fusion, hyper-parameters and multi-scale data volume are analyzed in depth.
The impact of phone-level supervision is experimentally excluded so the improvement obtained in the training stage should be attributed to the discriminative nature of the LF-MMI criterion. 
For results, state-of-the-art performances are achieved on two of the most popular Mandarin datasets: Aishell-1 and Aishell-2. In addition, considerable CER reduction is also achieved on a 30-hour low-resource ASR dataset and a 14.3k-hour industrial ASR dataset. 

This work also compares the proposed method with the MBR-based method. Experimentally we find the proposed LF-MMI method outperforms the MBR-based methods on the two E2E ASR frameworks. 
Also, the proposed training method achieves a much faster training speed than the MBR counterparts as the on-the-fly decoding process during training is reasonably eschewed. 
Finally, the proposed training method requires no pre-trained model for initialization but the MBR methods do. This sets us free from the complex training pipeline so the model can be trained in one go. 

For all, this work provides a new way to adopt LF-MMI, the criterion that was mainly adopted in the DNN-HMM system in previous research, into the E2E ASR systems. 
We believe it is feasible to use the proposed training and decoding method not only in the ASR research but also in practical systems.
In future work, we may explore the way to make the proposed method totally end-to-end so the adoption of phone lexicon can be eschewed.
We release all source code in the hope to make our work helpful to other researchers. 

\bibliographystyle{IEEEtran}
\bibliography{refs}

\section*{Appendix A}
All items in series $P(O_1^t|G*)$ can be obtained through the computation of the last term $P(O_1^T|G*)$ by recording the intermediate variables of the forward process. The forward computation is implemented frame-by-frame in logarithmic domain. Each state $i$ in graph (numerator or denominator) will have a state score $\alpha_t(i)$ (initialized by $\alpha_0(i)=-inf$) that will be updated in every frame step. The state scores are updated recursively until the last frame using the log-semiring:
\begin{equation}
    \alpha_{t+1}(i) = \log\sum_{a_{j,i}\in A(i)} exp(\alpha_{t}(j) + w_{t}(a_{j,i}))
\end{equation}
where $A(i)$ is the set of all arcs entering state $i$; $a_{j,i}$ is an arc from state $j$ to state $i$; $w_t(a_{j,i})$ is the predicted log-posterior for the input label on arc $a_{j,i}$ in $t$-th frame. Next, if we consider $t$-th frame as the last frame (a.k.a., to compute $\log P(O_1^t|G*)$), we should only consider the scores on the end states and add the state weights on those states. So, for any $1\le t\le T$, we have:
\begin{equation}
    \log P(O_1^t|G*) = \log \sum_{i}exp(s'_t(i))
\end{equation}
and
\begin{equation}
    s'_t(i) = \left\{
                 \begin{aligned}
                 & s_t(i) + \pi_i \quad \text{if state i is a final state}                        \\
                 & -\text{inf} \quad\quad \ \   \text{otherwise}                                \\
                 \end{aligned}
                 \right\}
\end{equation}
where $\pi_i$ is the weight on the final state $i$.

\end{document}